\documentclass{article} 
\usepackage{iclr2017_conference,times}
\usepackage{url}
\usepackage{color}
\usepackage{amsmath,amssymb}
\usepackage{soul}
\DeclareMathOperator{\E}{\mathbb{E}}

\title{Reinforcement Learning through Asynchronous Advantage Actor-Critic on a GPU}

\author{Mohammad Babaeizadeh \\ 
Department of Computer Science \\
University of Illinois at Urbana-Champaign, USA\\
\texttt{mb2@uiuc.edu}\\
\AND
Iuri Frosio, Stephen Tyree, Jason Clemons, Jan Kautz\\
NVIDIA, USA \\
\texttt{\{ifrosio,styree,jclemons,jkautz\}@nvidia.com}
}

\iclrfinalcopy 

\ifdefined\Arxive
	\newcommand{\ImgFolder}[1]{{#1}}%
\else
	\newcommand{\ImgFolder}[1]{Imgs/{#1}}%
\fi

\usepackage{subfig}			
\usepackage{graphicx}
\usepackage{enumitem}
\usepackage{wrapfig}
\usepackage{tabu}
\usepackage{relsize}
\usepackage{booktabs} 
\usepackage{multirow}
\usepackage{numprint}

\begin{document}

\maketitle

\begin{abstract}
We introduce a hybrid CPU/GPU version of the Asynchronous Advantage Actor-Critic (A3C) algorithm, currently the state-of-the-art method in reinforcement learning for various gaming tasks. We analyze its computational traits and concentrate on aspects critical to leveraging the GPU's computational power. We introduce a system of queues and a dynamic scheduling strategy, potentially helpful for other asynchronous algorithms as well. Our hybrid CPU/GPU version of A3C, based on TensorFlow, achieves a significant speed up compared to a CPU implementation; we make it publicly available to other researchers at \url{https://github.com/NVlabs/GA3C}.

\end{abstract}


\section{Introduction}
In the past, the need for task-specific, or even hand-crafted, features limited
the application of Reinforcement Learning (RL) in real world
problems~\citep{Sutton:1998:IRL:551283}. However, the introduction of Deep
Q-Learning Networks (DQN)~\citep{mnih-dqn-2015} revived the use of Deep Neural
Networks (DNNs) as function approximators for value and policy functions,
unleashing a rapid series of advancements. Remarkable results include learning
to play video games from raw
pixels~\citep{bellemare2016unifying,lample2016playing} and demonstrating
super-human performance on the ancient board game Go~\citep{44806}. Research
has yielded a variety of effective training formulations and DNN
architectures~\citep{HasseltGS15,WangFL15}, as well as methods to increase
parallelism while decreasing the computational cost and memory
footprint~\citep{NairSBAFMPSBPLM15,mnih-asyncrl-2016}.
In particular, \cite{mnih-asyncrl-2016} achieve state-of-the-art results on
many gaming tasks through a novel lightweight, parallel method called
Asynchronous Advantage Actor-Critic (A3C).
When the proper learning rate is used, A3C learns to play an Atari
game~\citep{OpenAIGym} from raw screen inputs more quickly and efficiently than previous methods:
on a $16$-core CPU, A3C achieves higher scores than previously published methods run for the same amount
of time on a GPU.



Our study sets aside many of the learning aspects of recent work and instead
delves into the computational issues of deep RL. Computational complexities are
numerous, largely centering on a common factor: RL has an inherently sequential
aspect, since the training data are generated while learning.
The DNN model is constantly queried to guide the actions of agents whose
gameplay in turn feeds DNN training. Training batches are commonly small and
must be efficiently shepherded from the agents and simulator to the DNN
trainer.
When using a GPU, the mix of small DNN architectures, small training batch
sizes, and contention for the GPU for both inference and
training can lead to a severe under-utilization of the computational resources.

To systematically investigate these issues, we implement both CPU and GPU
versions of A3C in TensorFlow (TF)~\citep{tensorflow2015-whitepaper},
optimizing each for efficient system utilization and to approximately replicate published scores
in the Atari 2600 environment~\citep{OpenAIGym}. We analyze a variety of ``knobs'' in the
system and demonstrate effective automatic tuning of those during training. Our hybrid
CPU/GPU implementation of A3C, named GA3C, generates and consumes training data substantially faster than its
CPU counterpart, up to $\sim\!6\times$ faster for small DNNs and $\sim\!45\times$
for larger DNNs. While we focus on the A3C architecture, this analysis
can be helpful for researchers and framework developers designing the next
generation of deep RL methods.

\section{Related work}

Recent advances in deep RL have derived from both novel algorithmic approaches and related systems
optimizations.
Investigation of the algorithmic space seems to be the most common approach
among researchers.
Deep Q-Learning Networks~(DQN) demonstrate a general approach to the learning
problem~\citep{mnih-dqn-2015}, relying heavily on the introduction of an
experience replay memory to stabilize the learning procedure.
This improves reliability but also increases the computational cost and memory footprint of the
algorithm. Inspired by DQN, researchers have proposed more effective
learning procedures, achieving faster and more stable convergence:
Prioritized DQN~\citep{SchaulQAS15} makes better use of the replay memory by
more frequently selecting frames associated with significant experiences.
Double-DQN~\citep{HasseltGS15} separates the estimate of the value function
from the choice of actions (policy), thus reducing the tendency in DQN to be
overly optimistic when evaluating its choices. Dueling Double
DQN~\citep{WangFL15} goes a step further by explicitly splitting the
computation of the value and advantage functions within the network. The presence of the replay memory makes the DQN approaches more suitable for a GPU implementation when compared to other LR methods, but state-of-the-art results are achieved by A3C~\citep{mnih-asyncrl-2016}, which does not make use of it.

Among systems approaches, AlphaGo~\citep{44806} recently achieved astonishing
results through combined algorithmic and hardware specialization.
The computational effort is impressive:
$40$ search threads, $1202$ CPUs, and $176$ GPUs are used in the distributed version for
inference only. Supervised training took around three weeks for the policy
network, using $50$ GPUs, and another day using the RL approach for refinement.
A similar amount of time was required to train the value network.
Gorilla DQN~\citep{NairSBAFMPSBPLM15} is a similarly impressive implementation
of distributed RL system, achieving a significant improvement over DQN. The
system requires $100$ concurrent actors on $31$ machines, $100$ learners and a
central parameter server with the network model. This work demonstrates the
potential scalability of deep RL algorithms, achieving better results in less
time, but with a significantly increased computational load, memory footprint, and cost.

\section{Asynchronous Advantage Actor Critic (A3C)}

\subsection{Reinforcement Learning Background}
In standard RL, an agent interacts with an environment over a number of discrete time steps. At each time step $t$, the agent observes a state $s_t$ and, in the discrete case, selects an action $a_t$ from the set of valid actions. An agent is guided by policy $\pi$, a function mapping from states $s_t$ to actions $a_t$. After each action, the agent observes the next state $s_{t+1}$ and receives feedback in the form of a reward $r_t$. This process continues until the agent reaches a terminal state or time limit, after which the environment is reset and a new episode is played.


The goal of learning is to find a policy $\pi$ that maximizes the expected reward. 
In policy-based model-free methods, a function approximator such as a neural network computes the policy $\pi(a_t|s_t;\theta)$, where $\theta$ is the set of parameters of the function. There are many methods for updating $\theta$ based on the rewards received from the environment. REINFORCE methods \citep{williams1992simple} use gradient ascent on $\E[R_t]$, where $R_t=\sum_{i=0}^{\infty}\gamma^ir_{t+i}$ is the  accumulated reward starting from time step $t$ and increasingly discounted at each subsequent step by factor $\gamma\in (0, 1]$.

The standard REINFORCE method updates $\theta$ using the gradient $\nabla_\theta\log\pi(a_t|s_t;\theta)R_t$, which is an unbiased estimator of $\nabla_\theta\E[R_t]$. The variance of the estimator is reduced by subtracting a learned \textit{baseline} (a function of the state $b_t(s_t)$) and using the gradient $\nabla_\theta\log\pi(a_t|s_t;\theta)\big(R_t~-~b_t(s_t)\big)$ instead. One common baseline is the value function defined as $V^\pi(s_t) = \E[{R_t|s_t}]$ which is the expected return for following the policy $\pi$ in state $s_t$.
In this approach the policy $\pi$ and the baseline $b_t$ can be viewed as \textit{actor} and \textit{critic} in an actor-critic architecture \citep{Sutton:1998:IRL:551283}.


\subsection{Asynchronous Advantage Actor Critic (A3C)}

\begin{figure}[t]
  \centering
  \subfloat[A3C]{\includegraphics[width=0.39\textwidth]{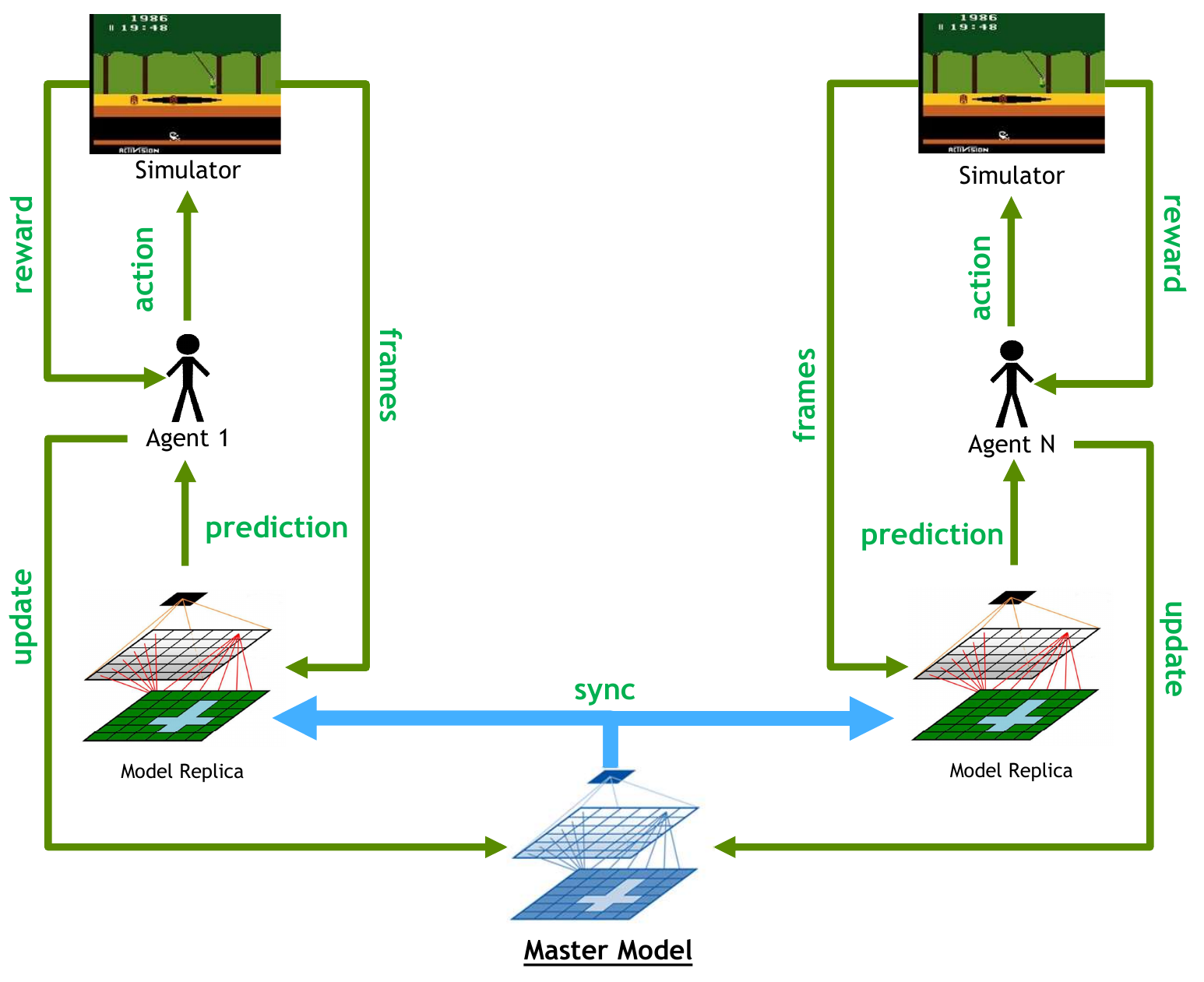} \label{fig:a3c}}
  \hfill
  \subfloat[GA3C]{\includegraphics[width=0.58\textwidth]{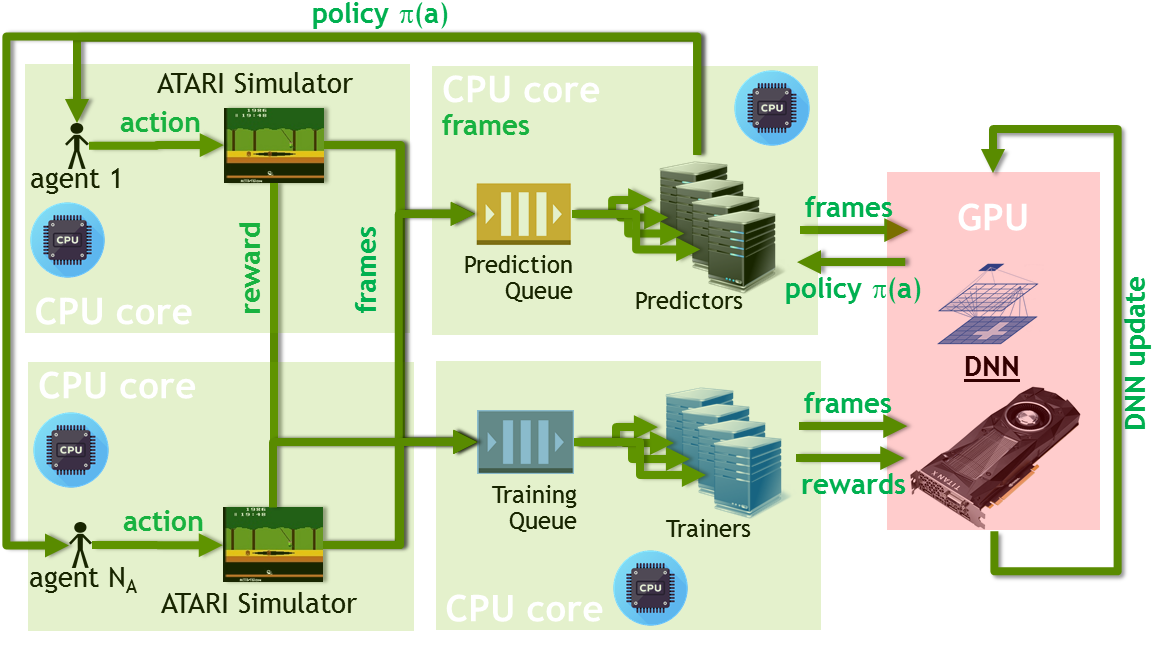} \label{fig:ga3c}}
  \caption{Comparison of A3C and GA3C architectures. Agents act concurrently both in A3C and GA3C. In A3C, however, each agent has a replica of the model, whereas in GA3C there is only one GPU instance of the model. In GA3C, agents utilize predictors to query the network for policies while trainers gather experiences for network updates.}
\end{figure}

A3C~\citep{mnih-asyncrl-2016}, which achieves state-of-the-art results on many gaming tasks including Atari 2600, uses a single DNN to approximate both the policy and value function. The DNN has two convolutional layers with 16$\times$8$\times$8 filters with a stride of 4, and 32$\times$4$\times$4 filters with a stride of 2, followed by a fully connected layer with 256 units; each hidden layer is followed by a rectifier nonlinearity. The two outputs are a softmax layer which approximates the policy function $\pi \left(a_t | s_t; \theta \right)$, and a linear layer to output an estimate of $V \left(s_t; \theta\right)$.
Multiple agents play concurrently and optimize the DNN through asynchronous gradient descent. Similar to other asynchronous methods, the network weights are stored in a central parameter server (Figure \ref{fig:a3c}). Agents calculate gradients and send updates to the server after every $t_{max} = 5$ actions, or when a terminal state is reached. After each update, the central server propagates new weights to the agents to guarantee they share a common policy.

Two cost functions are associated with the two DNN outputs. For the policy function, this is:
\begin{equation}
f_{\pi}\left( \theta \right)
 = \log\pi\left(a_t | s_t; \theta \right)
\left(R_t - V\left(s_t; \theta_t \right) \right)
+ \beta H \left(\pi \left(s_t; \theta \right)\right),
\label{eq:costPi}
\end{equation}
where $\theta_t$ are the values of the parameters $\theta$ at time $t$, $R_t = \sum_{i=0}^{k-1}{\gamma^i r_{t+i} + \gamma^k V\left(s_{t+k}; \theta_t \right)}$ is the estimated discounted reward in the time interval from $t$ to $t+k$ and $k$ is upper-bounded by $t_{max}$, while $H \left(\pi \left(s_t; \theta \right)\right)$ is an entropy term, used to favor exploration during the training process. The factor $\beta$ controls the strength of the entropy regularization term. The cost function for the estimated value function is:
\begin{equation}
f_v \left(\theta\right)
= \left( R_t - V \left( s_t; \theta \right) \right)^2.
\label{eq:costV}
\end{equation}

Training is performed by collecting the gradients $\nabla \theta$ from both of the cost functions and using the standard non-centered RMSProp algorithm \citep{tieleman2012lecture} as optimization:
\begin{equation}\
\begin{array}{l}
g = \alpha g + (1 - \alpha) \Delta \theta ^2\\
\theta \leftarrow \theta - \eta \Delta \theta / \sqrt{g + \epsilon}.
\end{array}
\label{eq:RMSProp}
\end{equation}
The gradients $g$ can be either shared or separated  between agent threads but the shared implementation is known to be more robust \citep{mnih-asyncrl-2016}. 

The original implementation of A3C~\citep{mnih-asyncrl-2016} uses $16$ agents on a $16$ core CPU and it takes about four days to learn how to play an Atari game~\citep{OpenAIGym}. The main reason for using CPU other than GPU, is the inherently sequential nature of RL in general, and A3C in particular. In RL, the training data are generated while learning, which means the training and inference batches are small and GPU is mostly idle during the training, waiting for new data to arrive. Since A3C does not utilize any replay memory, it is completely sequential and therefore a CPU implementation is as fast as a naive GPU implementation. 

\section{Hybrid CPU/GPU A3C (GA3C)}

We propose GA3C, an alternative architecture of A3C, with emphasize on an
efficient GPU utilization to increase the number of training data generated and processed per second. We demonstrate that our
implementation of GA3C effectively converges significantly faster than our CPU
implementation of A3C, achieving the state-of-the-art performance in a
shorter time.

\subsection{GA3C architecture}
The primary components of GA3C (Figure \ref{fig:ga3c}) are a DNN with training
and prediction on a GPU, as well as a multi-process, multi-thread CPU
architecture with the following components:
\begin{itemize}[style=unboxed,leftmargin=0cm]
	\item \textbf{Agent} is a process interacting with the simulation environment: choosing actions according to the learned policy and gathering experiences for further training.
Similar to A3C, multiple concurrent agents run independent instances of the
environment. Unlike the original, each agent does not have its own
copy of the model. Instead it queues policy requests in a \textit{Prediction
Queue} before each action, and periodically submits a batch of input/reward
experiences to a \textit{Training Queue}; the size of each training batch is typically equal to $t_{max}$ experiences, though it is sometimes smaller for experiences collected at the end of an episode.

	\item \textbf{Predictor} is a thread which dequeues as many prediction requests as are immediately available and batches them into a single inference query to the DNN model on the GPU.
When predictions are completed, the predictor returns the requested policy to
each respective waiting agent.
To hide latency, one or more predictors can act concurrently.

	\item \textbf{Trainer} is a thread which dequeues training batches submitted by agents 
and submits them to the GPU for model updates. GPU utilization can be increased by grouping training batches among several agents; we found that this generally leads to a more stable convergence, but the convergence speed is reduced when the merged training batches are too large; a compromise is explored in Section \ref{sec:policyLagResults}. Multiple trainers may run in parallel to hide latency.
\end{itemize}

Unlike A3C, GA3C maintains only one copy of the DNN model (Fig. \ref{fig:a3c} and \ref{fig:ga3c}), centralizing predictions and training updates and removing the need for synchronization. Also, in comparison with A3C, agents in GA3C do not compute the gradients themselves. Instead, they send experiences to trainers that update the network on the GPU accordingly. This introduces a potential lag between the generation and consumption of experiences, which we analyze in detail in Sections \ref{sec:policyLag} and \ref{sec:policyLagResults}.


\subsection{Performance Metrics and Trade-offs}
The GA3C architecture exposes numerous tradeoffs for tuning its computational
efficiency. In general, it is most efficient to transfer data to a GPU in large
enough blocks to maximize the usage of the bandwidth between the GPU and CPU.
Application performance on the GPU is optimized when the application has large
amounts of parallel computations that can hide the latency of fetching data
from memory. Thus, we want to maximize the parallel computations the GPU is
performing, maximize the size of data transfer to the GPU, and minimize the
number of transfers to the GPU. Increasing the number of predictors, $N_P$,
allows faster fetching prediction queries, but leads to smaller prediction
batches, resulting in multiple data transfers and overall lower GPU
utilization. A larger number of trainers, $N_T$, potentially leads to more
frequent updates to the model, but an overhead is paid when too many trainers
occupy the GPU while predictors cannot access it. Lastly, increasing the number
of agents, $N_A$, ideally generates more training experiences while hiding
prediction latency. However, we would expect diminishing returns from
unnecessary context switching overheads after exceeding some threshold
depending on the number of CPU cores.

These aspects are well captured by a metric like the \emph{Trainings Per Second} (TPS), which is the rate at which we remove batches from the training queue. It corresponds to the rate of model updates 
and it is approximately proportional to the overall learning speed, given a fixed
learning rate and training batch size.
Another metric is the
\emph{Predictions Per Second} (PPS), the rate of issuing prediction queries
from prediction queue, which maps to the combined rate of gameplay among all
agents.
Notice that in A3C a model update occurs every time an agent plays
$t_{max}\!=\!5$ actions~\citep{mnih-asyncrl-2016}. Hence, in a balanced
configuration, $\textrm{PPS}\!\approx\!\textrm{TPS}\!\times\!t_{max}$. Since
each action is repeated four times as in~\citep{mnih-asyncrl-2016}, the number
of frames per second is $4 \times$PPS.

Computational aspects are not disconnected from the convergence of the learning algorithm.
For instance, employing too many
agents will tend to fill the training queue, introducing a significant time delay
between agent experiences ($a_t$, $s_t$ and $R_t$ in Eq. (\ref{eq:costPi}))
and the corresponding model updates, possibly threatening model convergence
(see Section \ref{sec:policyLag}).
Another example is batching of
training data: larger batches improves GPU occupancy by increasing the parallelism. They also decrease the TPS (\emph{i.e.}, the number of model updates per second), increasing the chance that the DNN model used in prediction and to compute the gradient in Eq. (\ref{eq:costPiModDelay}) are indeed the same model. The consequence (experimentally observed, see Section \ref{sec:policyLagResults}) is an increased stability of the learning process but, beyond a certain training batch size, this leads to a reduction in the convergence speed.
In short, $N_T$, $N_P$, and $N_A$ encapsulate many complex dynamics relating
both computational and convergence aspects of the learning procedure. Their
effect on the convergence of the learning process has to be measured by
analyzing not only TPS but also the learning curves.

\subsection{Dynamic Adjustment of Trade-offs}
\begin{figure}
\center
  \includegraphics[width=\columnwidth]{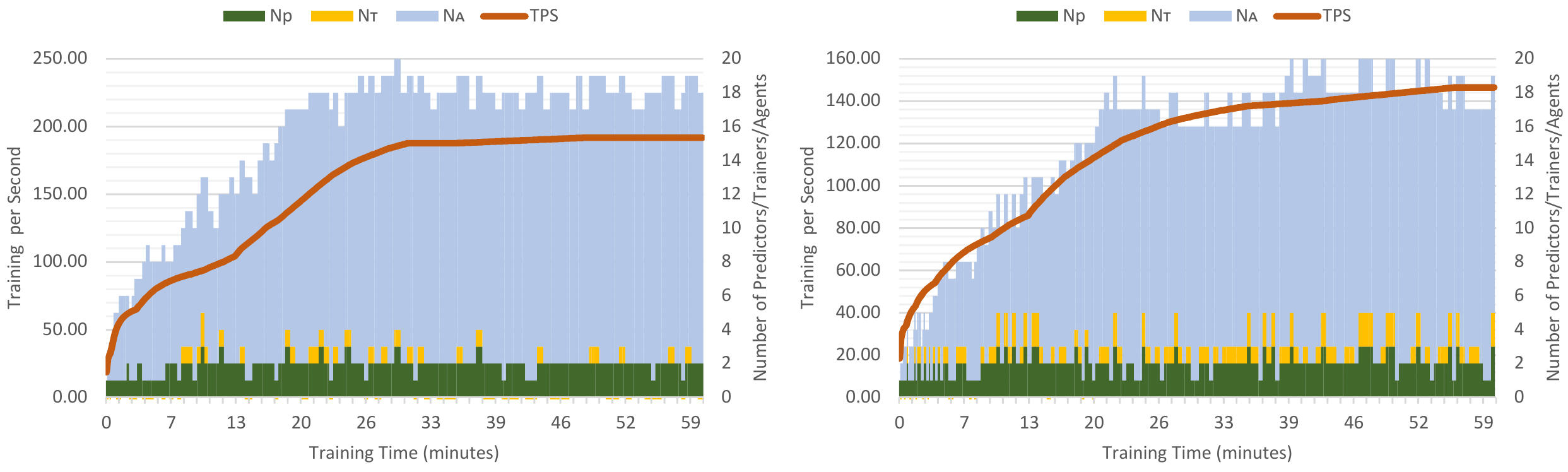}
  \caption{
  	Automatic dynamic adjustment of $N_T$, $N_P$, and $N_A$, to maximize TPS for {\sc Boxing} (left) and {\sc Pong} (right), starting from a sub-optimal configuration ($N_A\!=\!N_T\!=\!N_P\!=\!1$)
  }
  \label{fig:tps}
\end{figure}

The setting of $N_P$, $N_T$ and $N_A$ that maximizes the TPS depends on many aspects such as the
computational load of the simulation environment, the size of the DNN, and the
available hardware. As a rule of thumb, we
found that the number of agents $N_A$ should at least match the available CPU
cores, with two predictors and two trainers $N_P = N_T = 2$. However, this rule
hardly generalizes to a large variety of different situations and only
occasionally corresponds to the computationally most efficient configuration.
Therefore, we propose an annealing process to configure the system dynamically.
Every minute, we randomly change $N_P$, $N_T$, or $N_A$ by $\pm1$, monitoring
alterations in TPS to accept or reject the new setting. The optimal
configuration is then automatically identified in a reasonable time, for
different environments or systems. Figure~\ref{fig:tps} shows the automatic
adjustment procedure finding two different optimal settings for two different
games, on the same real system.

\subsection{Policy lag in GA3C}
\label{sec:policyLag} At a first sight, GA3C and A3C are different
implementations of the same algorithm, but GA3C has a subtle
difference which affects the stability of the algorithm. This problem is caused
by the latency between the time $t-k$, when a training example has been generated,
and when it is consumed for training, $t$, essentially changing the gradients
to:
\begin{equation}
\nabla_\theta\big[ \log\pi\left(a_{t-k} | s_{t-k}; \theta \right)
\left(R_{t-k} - V\left(s_{t-k}; \theta_t \right) \right)
+ \beta H \left(\pi \left(s_{t-k}; \theta \right)\right)\big].
\label{eq:costPiModDelay}
\end{equation}
Since the Training Queue is not blocking, the states it contains can be old. The value of the delay $k$ is bounded by the maximum size of the queue and influenced by how the system configuration balances training and prediction rates.
In other words, the DNN controller selects the action $a_{t-k}$ at time ${t-k}$; the corresponding experience lies in a training queue until time $t$, when a trainer thread pops the element out of the queue to compute the gradient as in Eq. (\ref{eq:costPiModDelay}). The DNN controller at time $t$ generally differs from the one at time $t-k$, since trainers can modify the DNN weights at any time. Therefore, the policy and value function $\pi$ and $V$ used to compute the gradient at time $t$ will differ from those used at time $t-k$ to collect the experience, whereas the action used to compute the gradient in Eq. (\ref{eq:costPiModDelay}) remains $a_{t-k}$.

This delay can lead to instabilities for two reasons. The first one is the possible generation of very large values in $\log\pi\left(a_{t-k} | s_{t-k}; \theta_{t} \right)$. In fact, $\pi\left(a_{t-k} | s_{t-k}; \theta_{t-k} \right)$ is generally large, since it is the probability of sampled action $a_{t-k}$, but over the course of lag $k$
new parameters $\theta_t$ can make $\pi\left(a_{t-k} | s_{t-k}; \theta_t \right)$ very small.
In the worst case, the updated probability is zero, generating infinite values in the $\log$ and causing optimization to fail. To avoid this, we add a small term $\epsilon>0$:
\begin{equation}
\nabla_\theta\big[ \log\big(\pi\left(a_{t-k} | s_{t-k}; \theta \right)+\epsilon\big)
\left(R_{t-k} - V\left(s_{t-k}; \theta_t \right) \right)
+ \beta H \left(\pi \left(s_{t-k}; \theta \right) + \epsilon \right)\big].
\label{eq:costPiEpsilon}
\end{equation}
Beyond fixing the error in the case $\pi=0$, this fix also improves the stability of the algorithm and removes the necessity of gradient clipping. In fact, as $\partial \log(\pi + \epsilon) / \partial \theta =  (\partial \pi / \partial \theta) / (\pi + \epsilon) $, $\epsilon$ establishes an upper bound for the multiplicative factor in front of $\partial \pi / \partial \theta$.
A similar term is also added in the entropy computation to avoid a similar explosion. It is important to remember that even A3C suffers from a similar issue. In fact, as the action $a_t$ in Eq. (\ref{eq:costPi}) is selected by sampling the output softmax, there is a chance that $\pi(a_t)$ is very small and therefore $\partial \log(\pi) / \partial \theta$ is large. However, gradient clipping prevents the usage of a gradient with large magnitude in A3C.

The second reason for introducing instabilities in GA3C is a generalization of the first one. Since training and predictions are computed with potentially different DNN parameters, the resulting gradient is noisy, and can therefore lead to unreliable updates of the DNN weights.
This is different from A3C, where every agent has its own copy of the model and uses it to compute both $\pi$ and $\partial \log(\pi) / \partial \theta$, before synchronizing the DNN model with the other agents.


\section{Analysis}


We profile the performance of GA3C and in the process seek to better understand
the system dynamics of deep RL training on hybrid CPU/GPU systems. Experiments
are conducted on the GPU-enabled systems described in
Table~\ref{tab:sys_prof_config} and monitored with CUDA profilers and custom
profiling code based on performance counter timing within Python.
We present profiling and convergence experiments both with and without
automatic adjustment of the number of agents $N_A$, trainers $N_T$, and
predictors $N_P$, and without constraints on the size of the prediction and
training queues.

\begin{figure}[h]
\centering
\includegraphics[trim={1cm 0cm 2cm 1cm},clip,width=0.8\textwidth]{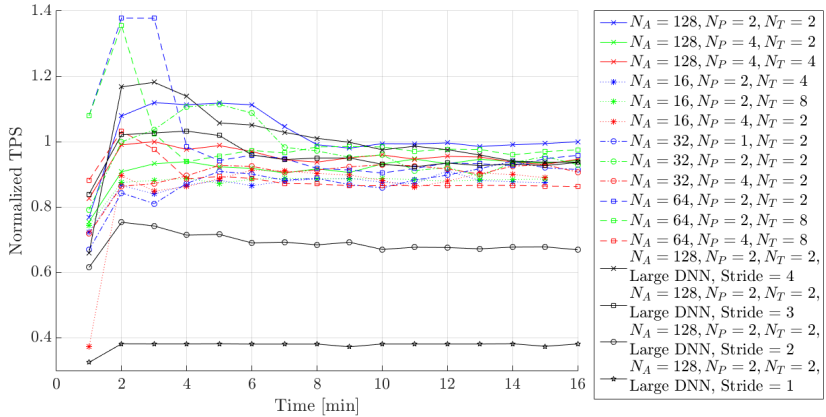}
\caption{TPS of the top three configurations of predictors $N_P$ and trainers $N_T$ for several settings of agents $N_A$, while learning {\sc pong} on System I from Table~\ref{tab:sys_prof_config}. TPS is normalized by best performance after $16$ minutes. Larger DNN models are also shown, as described in the text.}
\label{fig:UPS}
\end{figure}

\subsection{Effect of Resource Utilization on TPS}

\paragraph{Maximizing training speed.}
To begin, consider raw training speed as expressed in model update frequency,
or trainings per second (TPS). Figure~\ref{fig:UPS} shows TPS on System I in
Table~\ref{tab:sys_prof_config} for the first $16$ minutes of training on {\sc
pong}. We consider numbers of agents $N_A\!\in\!\{16,32,64,128\}$ and plot the
top $3$ combinations of $N_P,N_T\!\in\!\{1,2,4,8,16\}$. On this system,
increasing $N_A$ yields a higher TPS up to $N_A\!=\!128$ where diminishing
returns are observed, likely due to additional process overhead. The highest
consistent TPS on this system is observed with $N_A\!=\!128$ and
$N_P\!=\!N_T\!=\!2$ with a speed-up of $\sim4\times$ relative to the CPU-only
implementation (see Table \ref{tab:FPSs}).


\begin{table}[h]
\centering
\resizebox{0.8\textwidth}{!}{
\begin{tabular}{ccccc}
\toprule
& \textbf{System I} & \textbf{System II} & \textbf{System III} & \textbf{System IV}\\
\midrule
\multirow{3}{*}{\parbox{1.4cm}{\centering Processor (Intel)}} & Xeon E5-2640v3& Core i-3820 & Haswell E5-2698v3 & Xeon E5-2680v2\\
 & $2.60$ GHz & $3.60$ GHz & $2.30$ GHz & $2.80$ GHz\\
 & $16$ cores, dual socket & $8$ cores & $16$ cores & $10$ cores\\
\midrule
\multirow{2}{*}{\parbox{1.4cm}{\centering GPU (NVIDIA)}} & Geforce Titan X & GeForce 980 & Tesla K80 & Quadro M6000\\
 & (Maxwell) & (Maxwell) & (Kepler) & (Maxwell)\\
\midrule
Software / & \multicolumn{4}{c}{Python 3.5, CUDA 7.5 (I-III)/CUDA 8 (IV), CUDNN v5.1, TensorFlow r0.11}\\
Profilers & \multicolumn{4}{c}{nvprof, nvvp}\\
\bottomrule
\end{tabular}
}
\caption{Systems used for profiling and testing.}
\label{tab:sys_prof_config}
\end{table}


%

 
\begin{wraptable}{r}{0.60\textwidth}
\centering
\vspace{-0.75em}
\resizebox{0.60\textwidth}{!}{
\begin{tabular}{cccccccc}
\toprule
&& \multicolumn{3}{c}{\textbf{PPS}} && \multicolumn{2}{c}{\textbf{Utilization} ($\%$)}\\
\cmidrule{3-5}\cmidrule{7-8}
\bf{System} & \bf{DNN} & \bf{A3C} & \bf{GA3C} & \bf{Speed up} && \bf{CPU} & \bf{GPU}\\
\midrule
\multirow{5}{*}{{System I}} & small & $352$ & $1361$ & $4\times$ && $32$ & $56$\\
& large, stride $4$ & $113$ & $1271$ & $11\times$ && $27$ & $68$ \\
& large, stride $3$ & $97$ & $1206$ & $12\times$ && $27$ & $77$\\
& large, stride $2$ & $43$ & $874$ & $20\times$ && $26$ & $82$\\
& large, stride $1$ & $11$ & $490$ & $45\times$ && $17$ & $90$ \\
\midrule
\multirow{2}{*}{{System II}} & small & $116$ & $728$ & $6\times$ && $62$ & $33$\\
 & large, stride $1$ & $12$ & $336$ & $28\times$ && $49$ & $78$\\
\midrule
\multirow{2}{*}{{System III}} & small & $300$ & $1248$ & $4\times$ && $31$ & $60$\\ 
& large, stride $1$ & $38$ & $256$ & $6\times$ &&$17$ & $82$\\
\bottomrule
\end{tabular}
}
\caption{PPS on different systems (Table~\ref{tab:sys_prof_config}), for small and large DNNs, with CPU and GPU utilization for GA3C.
}
\label{tab:FPSs}
\vspace{-0.5em}
\end{wraptable}

%

\begin{figure}
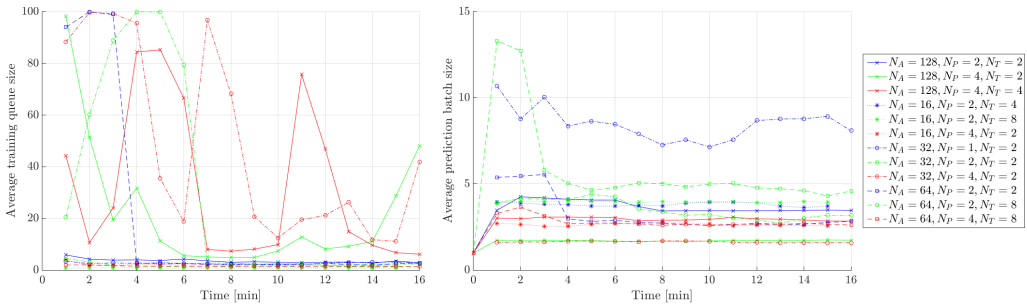

\centering
\subfloat{\includegraphics[trim={1cm 0cm 11.5cm 1cm},clip, width=0.41\textwidth]{\ImgFolder{AverageTrainingQueueSize_EachMinute.png}}}
\subfloat{\includegraphics[trim={1cm 0cm 1.2cm 1cm},clip, width=0.59\textwidth]{\ImgFolder{AveragePredictionBatchSize_EachMinute.png}}}
\caption{The average training queue size (left) and prediction batch size (right) of the top $3$ performing configurations of $N_P$ and $N_T$, for each $N_A$, with {\sc pong} and the System I in Table~\ref{tab:sys_prof_config}.}
\label{fig:AveQSizeAndBatch}
\end{figure}



\paragraph{GPU utilization and DNN size.}
The fastest configuration ($N_A\!=\!128$, $N_P\!=\!N_T\!=\!2$) for System~I in
Table \ref{tab:sys_prof_config} has an average GPU utilization time of only
$56\%$, with average and peak occupancy of $76\%$ and $98\%$,
respectively.\footnote{Occupancy reflects the parallelism of the computation by
capturing the proportion of GPU threads which are active while the GPU is in
use. For example, occupancy of $76\%$ with utilization of $56\%$ implies
roughly $43\%$ overall thread utilization.} This suggests there is
computational capacity for a larger network model. Therefore we profile GA3C on
a series of deeper DNN architectures\footnote{We increase the number of filters
in the first layer of the DNN from $16$ to $32$, and add a third convolutional
layer with $64$ $4\times4$ filters and stride $2$. To generate four large DNNs
with varying computational cost, we reduce the stride on the first layer from
$4$ pixels to $1$ pixel, with a step of $1$ pixel. Note the DNN with stride $1$
in the first layer also has $4\times$ larger fully-connected layer.} to
evaluate this hypothesis. Figure~\ref{fig:UPS} shows that TPS drops by only
$7\%$ with a one-layer deeper DNN controller; at the same time, the average GPU
utilization and occupancy increase by approximately $12\%$ and $0.5\%$,
respectively. The $7\%$ drop in TPS is the consequence of the increased depth
which forces an additional serial computational on the GPU (and therefore a
$12\%$ increase in its utilization). The negligible $0.5\%$ increase in
occupancy is likely explained by an efficient management of the computational
resources by cuDNN; there is still room available to run additional parallel
tasks (or, in other words, a wider DNN) at minimal cost.

By reducing the stride of the first layer of the DNN (in addition to adding a
convolutional layer), we scale DNN size with finer granularity, and we compare
FPS between GA3C and our CPU implementation of A3C.  Table \ref{tab:FPSs} shows
the speed up provided by GA3C increases as the DNN grows. This is mainly due to
increasing GPU utilization, as reported in Table \ref{tab:FPSs}. With the
largest DNN, our CPU implementation achieves TPS $\approx 11$, which is
approximately $45\times$ slower than GA3C. This behavior is consistent across
different systems, as shown in Table \ref{tab:FPSs}, where the CPU
implementation of A3C using the largest DNN with stride $1$ is $7\times$
(System III) to $32\times$ (System I) slower than the small network. Scaling
with DNN size is more favorable on a GPU, with a slow down factor of
$4.9\times$ in the worst case (System III) and $2.2\times$ in the best case
(System II). Further, more recent GPUs (Maxwell architecture) scale better
($2.2\times$ and $2.7\times$ slow down for Systems I and II) than older GPUs
($4.8\times$ slow down for the Kepler architecture, System III).

Generally speaking, for large DNNs, maximum TPS and FPS are achieved by
intensively using the GPU for prediction and training, while the CPU runs the
simulation environment and remains mostly idle. In practice, this allows
experimenting with larger architectures, which may be particularly important
for real world problems, e.g. robotics or autonomous driving~\citep{Lil15}.
Moreover, the idle CPU represents an additional computational resource, but
such investigation is beyond the scope of this paper.


\paragraph{Significant latency.}
Profiling on System I in Table \ref{tab:sys_prof_config} reveals that the
average time spent by an agent waiting for a prediction call to be completed is
$108$ms, only $10\%$ of which is taken by the GPU inference. The remaining
$90\%$ is overhead spent accumulating the batch and calling the prediction
function in Python. Similarly, for training we find that of the average
$11.1$ms spent performing a DNN update, $59\%$ is overhead. This seems to
suggest that a more optimized implementation (possibly based on a low level
language like C++) may reduce these overheads, but this investigation remains
for future work. 

\paragraph{Manually balancing components.}
Agents, predictors, and trainers all share the GPU as a resource; thus balance
is important. Figure~\ref{fig:UPS} shows the top three performing
configurations of $N_P$ and $N_T$ for different numbers of agents, $N_A$, with
System I in Table \ref{tab:sys_prof_config}. A $14\%$ drop in TPS is exhibited
between the best and worst depicted configuration, despite the exclusion of all
but the top three performers for each number of agents. The best results have
$4$ or fewer predictor threads, seemingly preventing batches from becoming too
small. The $N_P\!:\!N_T$ ratios for top performers tend to be $1\!:\!2$,
$1\!:\!1$, or $2\!:\!1$, whereas higher ratios such as $1\!:\!8$ and $1\!:\!4$
are rarely successful, likely due to the implicit dependence of training on
prediction speed. However, if the training queue is too full, training calls
take more GPU time, thereby throttling prediction speed. This is further
confirmed by our experimental finding that TPS and PPS plots track closely.
Figure~\ref{fig:AveQSizeAndBatch} shows training queue size and prediction
batch size for the top configurations. In all cases, the training queue
stabilizes well below its maximum capacity.
Additionally, the fastest configuration has one of the largest average
prediction batch sizes, yielding higher GPU  utilization.

\begin{figure}[tb]
	\vspace{-1em}
  \captionsetup[subfigure]{labelformat=empty}
  \centering
  \includegraphics[width=\textwidth]{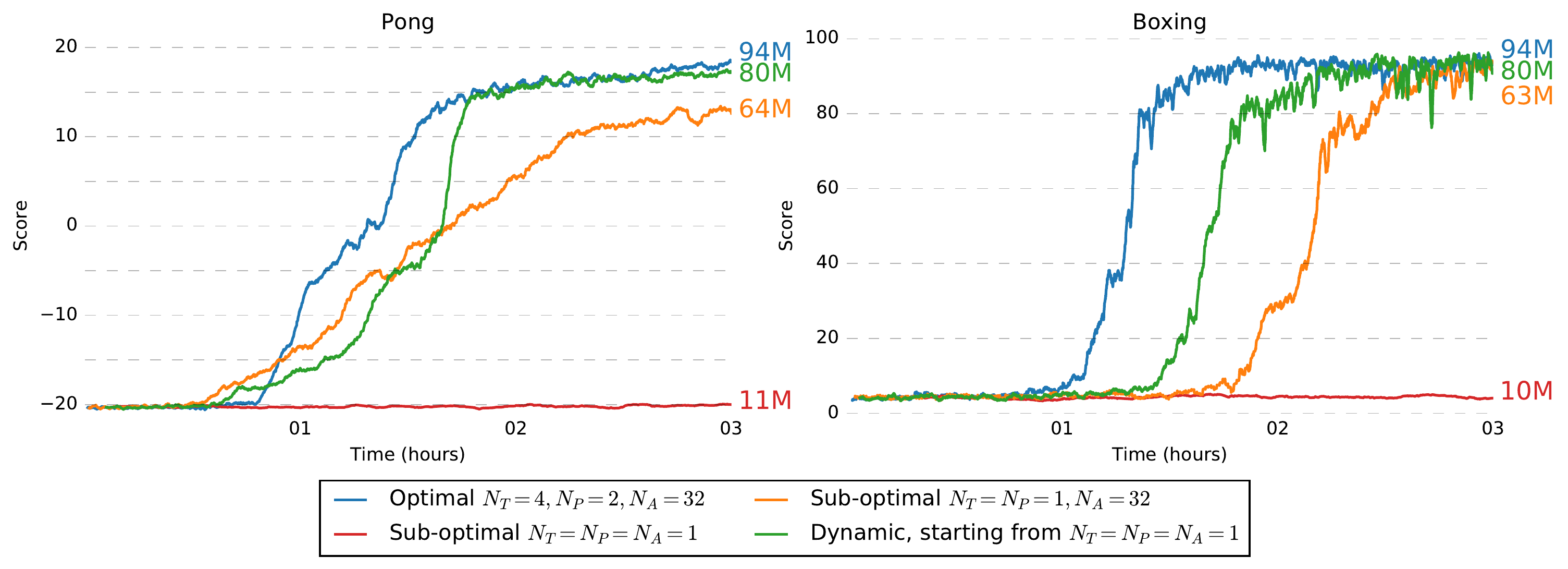}
  \caption{Effect of PPS on convergence speed. For each game, four different settings of GA3C are shown, all starting from the same DNN initialization. Numbers on the right show the cumulative number of frames played among all agents for each setting over the course of $3$ hours. Configurations playing more frames converge faster. The dynamic configuration method is capable of catching up with the optimal configuration despite starting with a sub-optimal setting, $N_T\!=\!N_P\!=\!N_A\!=\!1$.}
  \label{fig:DynamicVsNonDynamicConvergence}
  \vspace{-0.5em}
\end{figure}

\subsection{Effect of TPS on Learning Speed}

The beneficial effect of an efficient configuration on the training speed is
shown in Figure~\ref{fig:DynamicVsNonDynamicConvergence}. Training with a
suboptimal configuration (e.g. $N_P\!=\!N_T\!=\!N_A\!=\!1$ or
$N_P\!=\!N_T\!=\!1$, $N_A\!=\!16$) leads to a severe underutilization of the
GPU, a low TPS, and a slow training process. Using the optimal configuration
achieves a much higher score in a shorter period of time, mainly driven by
playing more frames, i.e. collecting more experiences, in the same amount of
time.

\citet{mnih-asyncrl-2016} note that asynchronous methods generally achieve
significant speedups from using a greater number of agents, and even report
superlinear speedups for asynchronous one-step Q-learning.
It is worth noting that optimal configurations for GA3C generally employ a much
higher number of agents compared to the CPU counterpart, e.g. the optimal
configuration for System I in Table \ref{tab:sys_prof_config} uses $128$
agents. This suggests that GPU implementations of asynchronous learning methods
may benefit from both a higher TPS and from collecting experiences from a wider
number of agents.

The learning curve for GA3C with dynamic configuration
(Figure~\ref{fig:DynamicVsNonDynamicConvergence}) tracks closely with the
learning curve of the optimal configuration. The total number of frames played
is generally slightly lower over the same time due to the search procedure overhead: the
configuration is changed once every minute, tending to oscillate around the
optimal configuration. Notice also that, in Figure~\ref{fig:DynamicVsNonDynamicConvergence}, the starting point of the dynamic configuration is $N_T = N_P = N_A = 1$, which is much slower than the optimal configuration. But scoring performance is nearly identical, indicating
that the dynamic method may ease the burden of configuring GA3C on a new
system.

Table~\ref{tab:scores} compares scores achieved by A3C on the CPU (as reported
in~\citep{mnih-asyncrl-2016}) with the best agent trained by our
TensorFlow implementation of GA3C. Unfortunately, a direct speed comparison is
infeasible without either the original source code or the average number of
frames or training updates per second.
However, results in this table do show that after one day of training our
open-source implementation can achieve similar scores to A3C after four days of
training.

\begin{figure}[t]
\includegraphics[trim={0cm 0cm 0cm 0cm},clip,width=\textwidth]{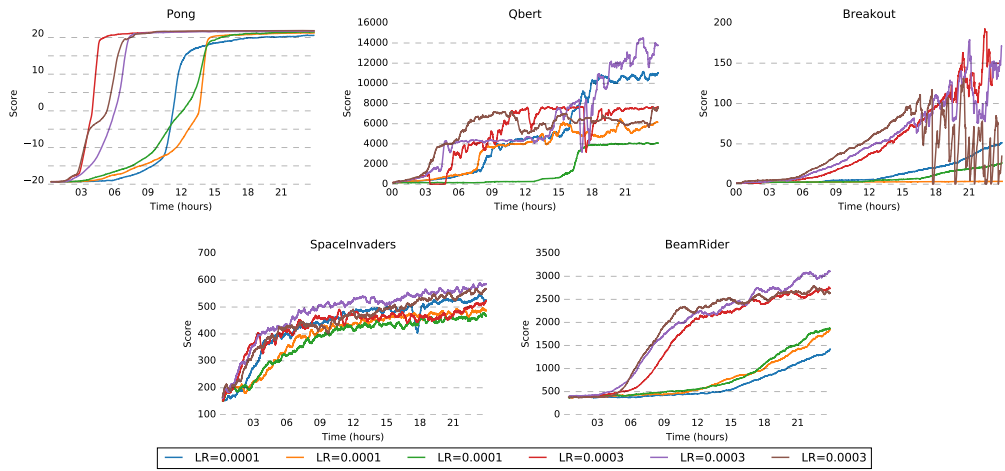}  \caption{Training curves for GA3C on five Atari games. Each training has been performed three times for each of two learning rates ($0.0003$ and $0.0001$) on System IV in Table~\ref{tab:sys_prof_config}.}
\label{fig:trainingCurves}
\end{figure}

Figure~\ref{fig:trainingCurves} shows typical training curves for GA3C on
several Atari games as a function of wall-clock time. When compared to the training curves reported in \cite{mnih-asyncrl-2016}, GA3C shows faster convergence toward the maximum
score in a shorter time for certain games such as {\sc Pong}, convergence
towards a better score in a larger amount of time (e.g. {\sc QBert}) or, for
other games, a slower convergence rate (e.g. {\sc Breakout}). It has to be
noted, however, that data reported by \cite{mnih-asyncrl-2016} are the average
learning curves of the top five learners in a set of fifty learners, each with
a different learning rate. On the other hand, in Figure ~\ref{fig:trainingCurves}, we are reporting three different runs for two (not essentially optimal) learning rates, fixed for all the games. This demonstrates some robustness of GA3C with respect to the choice of the learning rate, whereas it is also likely that better learning curves can be obtained using optimized learning rates. A deeper investigation on a large amount of data, potentially facilitated by our release of the GA3C code, may also reveal how peculiarities of each game differently affect the convergence of A3C and GA3C,
but this goes beyond the scope of this paper.
%


\newcommand{\game}[1]{{\footnotesize\sc\lowercase{#1}}} 
\newcommand{\stat}[1]{\footnotesize{#1}} 
\newcommand{\smallsim}{\mathsmaller{\sim}}
\newcommand{\n}[1]{$\numprint{#1}$}
\nprounddigits{0} \npdecimalsign{.} \npthousandsep{}
\newcommand{\hide}[1]{}

\begingroup\tabcolsep=2pt
\begin{table}
\begin{center}
\resizebox{\linewidth}{!}{
\begin{tabular}{rrrrrrrrrrrrccc}
\toprule
&& \multicolumn{9}{c}{\textbf{Atari Game Scores}} && \multicolumn{2}{c}{\textbf{Attributes}} \\
\cmidrule{3-11}\cmidrule{13-14}
&& \game{Amidar} & \game{Boxing} & \game{Centipede} & \game{\tiny Name This Game} & \game{Pacman} & \game{Pong} & \game{Qbert} & \game{Seaquest} & \game{Up-Down} && \stat{Time} & \stat{System} \\
\midrule
 \textbf{Human} && \n{1675.8} & \n{9.6}  & \n{10321.9} & \n{6796.0}  & \n{15375.0} & \n{15.5}  & \n{12085.0} & \n{40425.8} & \n{9896.1}  && --       & --  \\
 \textbf{Random} && \n{5.8}&\n{-1.5}&\n{1925.5}&\n{197.8}&\n{1747.8}&\n{-18.0}&\n{271.8}&\n{215.5} & \n{533.4} && -- & -- \\
 \textbf{A3C} && \n{263.9}  & \n{59.8} & \n{3755.8}  & \n{10476.1} & \n{653.7}   & \n{5.6}   & \n{15148.8} & \n{2355.4}  & \n{74705.7} && $4$ days & CPU \\
  \textbf{GA3C}                          && \n{217.7}  & \n{91.8} & \n{7386.1}  & \n{5642.6}  & \n{1978.3}  & \n{18.1}  & \n{14965.8} & \n{1706.0}  & \n{8623.3}  && $1$ day  & GPU \\
\bottomrule
\end{tabular}
}
\end{center}
\vspace{0.25em}
\caption{Average scores on a subset of Atari games achieved by:
a random player~\citep{mnih-dqn-2015};
a human player~\citep{mnih-dqn-2015};
A3C after four days of training on a CPU~\citep{mnih-asyncrl-2016};
and GA3C after one day of training. For GA3C, we measured the average score on 30 games, each initialized with a random seed.}
\label{tab:scores}
\end{table}
\npnoround
\endgroup

\npnoround

\begin{figure}
\includegraphics[trim={0cm 0cm 0cm 0cm},clip,width=\textwidth]{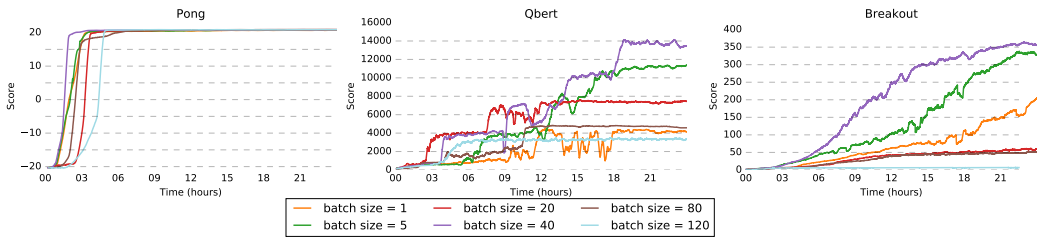}
\caption{Training GA3C with a range of minimum training batch sizes. Increasing the minimum training batch size from $1$ to $40$ reduces the effect of the policy lag (delay $k$ in Eq.~(\ref{eq:costPiModDelay})), leading to convergence that is faster and more stable. GA3C achieved the overall best results with a minimum batch size between $20$ and $40$. Increasing beyond this threshold dramatically reduces convergence speed for some games, especially those inclined to unstable learning curves.}
\label{fig:trainingDifferentBatchSizes}
\end{figure}


\subsection{Policy lag, learning stability and convergence speed}
\label{sec:policyLagResults}

One of the main differences between GA3C and A3C is the asynchronous computation
of the forward step (policy $\pi$) and the gradients (Eq. (\ref{eq:costPiModDelay})) used to update the DNN. Delays between these two operations may introduce noise in the
gradients, making the learning process unstable. We experimentally
investigated the impact of this asynchrony on the learning
process to determine if a synchronization mechanism, which may
negatively impact both PPS and TPS, can increase the stability of the algorithm.



In GA3C, each agent generally pushes $t_{max}$ experiences in the training queue.
By default, trainers collect a batch of experiences from a single agent from the training queue and send the batch to the GPU to compute gradients, as in Eq. (\ref{eq:costPiEpsilon}).
Each time a trainer updates the DNN weights, the remaining experiences in the training queue are no longer in sync with the DNN model. This situation becomes worse when the average length of the training queue is large.

By allowing larger training batch sizes, we reduce the number of DNN updates per second (TPS), and consequently diminish the effect of the delay $k$ in Eq. (\ref{eq:costPiEpsilon}). In this way we increase the chance that the collected experiences and the computed gradients are in sync, which improves the stability.
Notice that, even if the TPS is lower, the average magnitude of the updates is indeed larger, since we sum the gradients computed over the training batch.

In this setting, the optimal training batch size compromises among TPS, the average gradient step magnitude and the training stability. Another factor to be considered is that batching training data potentially leverages the GPU computational capability better by reducing the time devoted to compute the DNN updates while increasing the GPU occupancy during this phase. This gives more GPU time to the predictors, potentially increasing the PPS. However, this advantage tends to disappear when the training batch size is too large and predictors stay idle while the DNN update is computed.

Figure~\ref{fig:trainingDifferentBatchSizes} compares convergence curves when
no minimum size for training batch is compulsory (the default GA3C
implementation where gradient updates are computed on a single agent's batch)
and when a minimum training batch size is enforced (combining multiple agent
batches into a single gradient update). In the latter case, trainers collect
experiences from multiple agents at the same time from the training queue and
send them to the GPU for computation of gradients as in Eq.
(\ref{eq:costPiEpsilon}). Up to a certain batch size (between $20$ and $40$, in
our experiments), increasing the training batch size stabilizes the learning
procedure and generally leads to faster convergence. Some games such as {\sc
Pong} indeed do not suffer from this instability, and the effect of the minimum
batch size is less evident in this case. We speculate that a careful selection
of the learning rate combined with the proper minimum training batch size may
lead to even faster convergence.

\section{Conclusion}


By investigating the computational aspects of our hybrid CPU/GPU implementation
of GA3C, we achieve a significant speed up with respect to its CPU counter
part. This comes as a result of a flexible system capable of finding a
reasonable allocation of the available computational resources. Our approach
allows producing and consuming training data at the maximum pace on different
systems, or to adapt to temporal changes of the computational load on one
system. Despite the fact that we analyze A3C only, most of our findings can be
applied to similar RL asynchronous algorithms.

We believe that the analysis of the computational aspects of RL algorithms may be a consistent theme in RL in the future, motivating further studies such as this one. 
The potential benefits of such investigation goes well beyond the computational
aspects. For instance, we demonstrate that GA3C scales with the size of the DNN
much more efficiently than our CPU implementation of A3C, thus opening the
possibility to explore the use of large DNN controllers to solve real world RL
problems.

By open sourcing GA3C (see \url{https://github.com/NVlabs/GA3C}), we allow other researchers to further explore this
space,
investigate in detail the computational aspects of deep RL algorithms, and test
new algorithmic solutions, including strategies for the combined utilization of
the CPU and GPU computational resources.



\subsubsection*{Acknowledgments}

We thank Prof. Roy H. Campbell for partially supporting this work.

\bibliography{references}
\bibliographystyle{iclr2017_conference}

\end{document}